\documentclass[review]{elsarticle}

\usepackage[colorinlistoftodos]{todonotes}
\usepackage{multirow}
\usepackage{rotating}
\usepackage{lineno,hyperref}
\usepackage{textgreek}
\modulolinenumbers[5]
\usepackage{xcolor}
\usepackage{amsmath}
\usepackage{amsfonts}
\usepackage{algorithm}
\usepackage{caption}
\usepackage{algpseudocode}
\usepackage{tabu}
\newcolumntype{M}[1]{>{\centering\arraybackslash}m{#1}}
\newcolumntype{L}[1]{>{\arraybackslash}m{#1}}
\newcolumntype{R}[1]{>{\raggedleft\let\newline\\\arraybackslash\hspace{0pt}}m{#1}}
\journal{Journal of Biomedical Informatics}

\begin{document}

\begin{frontmatter}

\title{Large-scale investigation of weakly-supervised deep learning for the fine-grained semantic indexing of biomedical literature}

\fntext[myfootnote]{This is a preprint of an article published in the Journal of Biomedical Informatics. The final authenticated version is available online at: https://doi.org/10.1016/j.jbi.2023.104499}

\author[NCSRD,AUTH]{Anastasios Nentidis\corref{mycorrespondingauthor}}
\cortext[mycorrespondingauthor]{Corresponding author}
\ead{nentidis@csd.auth.gr, tasosnent@iit.demokritos.gr }
\author[NCSRD,upatras]{Thomas Chatzopoulos}
\author[NCSRD]{Anastasia Krithara}
\author[AUTH]{Grigorios Tsoumakas}
\author[NCSRD]{Georgios Paliouras}


\address[NCSRD]{Institute of Informatics and Telecommunications, NCSR Demokritos, Athens, Greece}
\address[AUTH]{School of Informatics, Aristotle University of Thessaloniki, Thessaloniki, Greece}
\address[upatras]{Department of Computer Engineering and Informatics, University of Patras, Patras, Greece}

\begin{abstract}
Objective: Semantic indexing of biomedical literature is usually done at the level of MeSH descriptors with several related but distinct biomedical concepts often grouped together and treated as a single topic. This study proposes a new method for the automated refinement of subject annotations at the level of MeSH concepts. 
Methods: Lacking labelled data, we rely on weak supervision based on concept occurrence in the abstract of an article, which is also enhanced by dictionary-based heuristics. 
In addition, we investigate deep learning approaches, making design choices to tackle the particular challenges of this task. 
The new method is evaluated on a large-scale retrospective scenario, based on concepts that have been promoted to descriptors.   
Results: In our experiments concept occurrence was the strongest heuristic achieving a macro-F1 score of about 0.63 across several labels. 
The proposed method improved it further by more than 4pp. 
Conclusion: The results suggest that concept occurrence is a strong heuristic for refining the coarse-grained labels at the level of MeSH concepts and the proposed method improves it further.  
\end{abstract}


\end{frontmatter}


\section{Introduction}

Biomedical literature is a rich source of biomedical knowledge that is constantly expanding. 
Identification of relevant articles for specific topics in such a rich collection is a real challenge.
In this direction, the National Library of Medicine (NLM) indexes citations in the MEDLINE database with topic descriptors from the Medical Subject Headings (MeSH) thesaurus\footnote{\url{https://meshb.nlm.nih.gov/}}. 
This semantic indexing process allows MEDLINE/PubMed\footnote{\url{https://pubmed.ncbi.nlm.nih.gov/}} to offer advanced semantic search strategies, that can retrieve citations relevant to a topic of interest, addressing issues such as synonymy and polysemy of biomedical concepts.

\begin{figure}[!t]
\centerline{\includegraphics[width=0.7\textwidth]{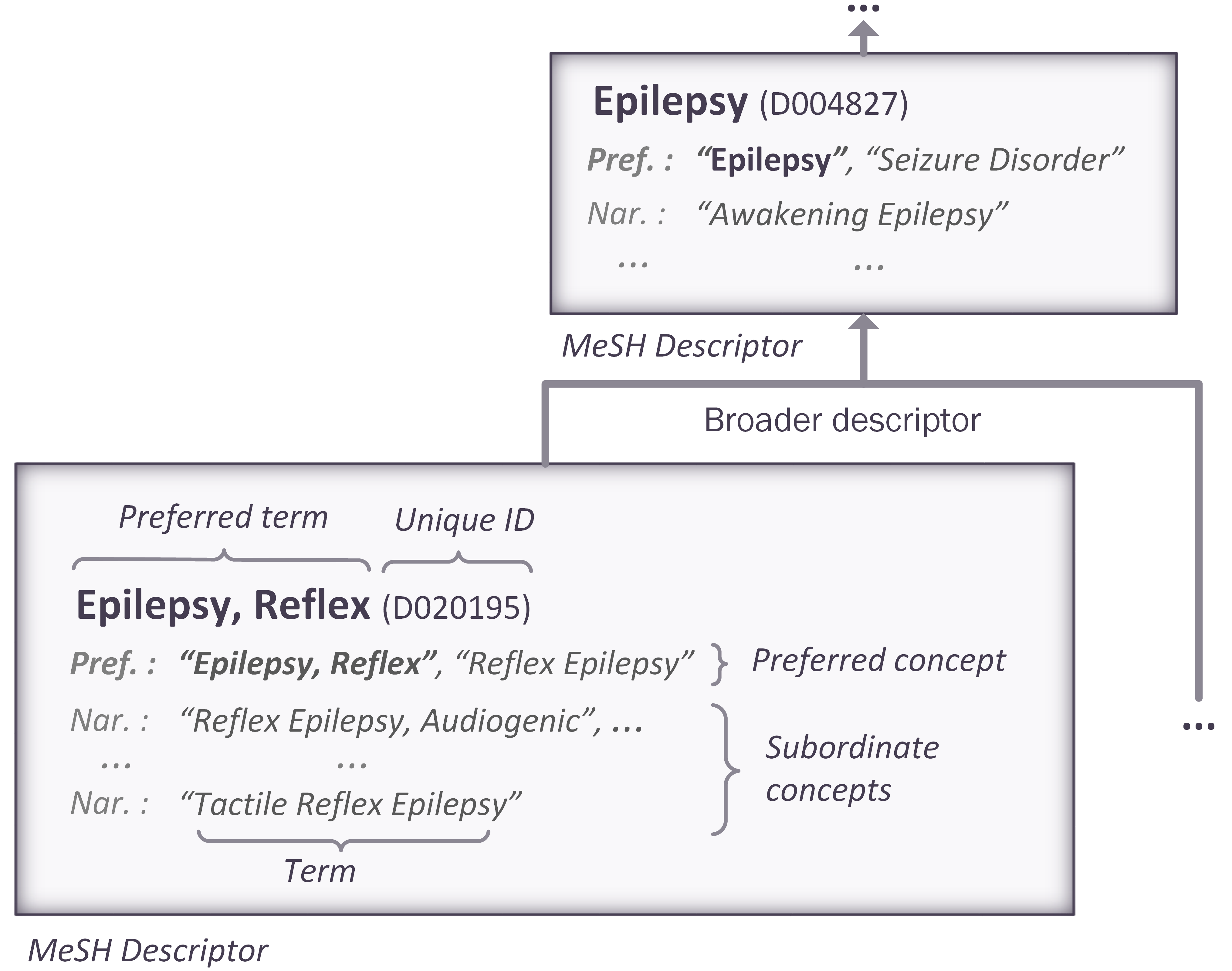}}
\caption{ The structure and hierarchical position of the MeSH descriptor ``Epilepsy, Reflex''. }\label{fig:MeSH_Example}
\end{figure}

MeSH provides more than thirty thousand descriptors, such as diseases and chemicals, hierarchically organized so that most are subcategories of at least one broader descriptor.
For example, ``Epilepsy, Reflex''
is a subcategory of the broader descriptor ``Epilepsy''
as shown in Figure~\ref{fig:MeSH_Example}.
Some of these descriptors cover several related but distinct subordinate \textit{MeSH concepts}\footnote{\url{https://www.nlm.nih.gov/mesh/concept_structure.html}}, which are defined as sets of synonymous terms. 
These concepts are usually narrower\footnote{In some cases subordinate concepts can be broader than the main topic.} than the main topic of the descriptor. 
For example, ``Epilepsy, Reflex'' contains a \textit{preferred concept}, providing the name of the descriptor and the synonymous term ``Reflex Epilepsy'', but includes also additional terms for narrower concepts such as ``Reflex Epilepsy, Audiogenic'' and ``Tactile Reflex Epilepsy'', that are not MeSH descriptors themselves.

Such narrower concepts are not distinguishable by the semantic indexing process, which is done at the level of descriptors. 
Therefore, semantic search cannot be used at this level of detail. 
The goal of this work is to enable such fine-grained semantic search, providing a method for the Fine-Grained Semantic Indexing ($FGSI$)
of the biomedical literature, that is the automated refinement of MeSH labels from the descriptor level to the level of each narrower concept. 
For example, we would like to distinguish which of the articles already annotated with the ``Epilepsy, Reflex'' are for ``Audiogenic Epilepsy'' and which for ``Tactile Reflex Epilepsy'' or other subcategories of ``Epilepsy, Reflex''.

Indexing at the level of MeSH concepts is beyond the current practice in MEDLINE/PubMed and, if enforced, it would almost double the number of distinct topics used for indexing.
Doing this additional work manually is not practical, given the growing volumes of biomedical literature.
Furthermore, state-of-the-art systems that provide automated topic suggestions 
focus on known MeSH topics and usually require several labelled training examples for each topic~\cite{Nentidis2022}. Therefore, obtaining reliable topic suggestions for new fine-grained topics at the level of concepts is an open research issue. 
The lack of labelled data for the development and evaluation of methods is a major challenge of this $FGSI$ task.

In this work, we propose \textit{Deep Beyond MeSH} ($DBM$), a new method for the refinement of coarse-grained MeSH annotations based on deep pre-trained language models, namely \textit{PubMedBERT}~\cite{Gu2022}. 
To overcome the lack of labelled data for supervised training, $DBM$ relies on weak supervision, investigating whether such  Deep Learning (DL) models can be resilient to heuristic labels.   
For weak supervision, the method is based on \textit{concept occurrence} ($CO$)\footnote{This is based on concept recognition by the \textit{MetaMap} tool~\cite{Aronson2004}.}, which is defined as the information extraction task of recognizing a biomedical named entity in a text and then mapping it to a specific concept from a normalized semantic system, such as the \textit{UMLS}\footnote{\url{https://www.nlm.nih.gov/research/umls/index.html}}. 

Previous experiments with this idea~\cite{nentidis2020beyond}, using small ground-truth datasets for refining two specific MeSH descriptors, suggested that $CO$ can provide a strong basis for the $FGSI$ task. 
However, this heuristic needed to be tested on larger datasets of ground-truth data that cover several MeSH descriptors. 
In this direction, we devised a retrospective scenario, based on the evolution of MeSH indexing in MEDLINE, and introduce here \textit{Retrospective Beyond MeSH} ($RetroBM$). This new method derives large-scale datasets for several individual concepts that got promoted over time to new fine-grained MeSH descriptors, that is descriptors covering a single MeSH concept. 

In brief, the main questions driving this work are the following:
\begin{itemize}
    \item Is $CO$ a good heuristic for concept-level $FGSI$ at a large scale? 
    Could $CO$ be enhanced when combined with other heuristics?
    \item Is it possible to train DL models for $FGSI$ without ground-truth training data, by using the above heuristics as weak supervision? 
    Could these models exceed the predictive performance of the heuristics themselves?
    \item Can the natural extension of MeSH towards fine-grained labels be exploited for the retrospective evaluation of automated $FGSI$ approaches at a large scale?
\end{itemize}

The rest of this paper is organized as follows.
In Section~\ref{sec:Background}, we provide some background and previous efforts towards $FGSI$ of biomedical literature.
In Section~\ref{sec:Methods}, we introduce two new methods: $RetroBM$ for the development of evaluation datasets and $DBM$ for indexing documents with fine-grained labels without ground-truth data.
In Section~\ref{sec:Experiments}, we present the experiments and the results of this work.
Finally, in Section~\ref{sec:Conclusion} we summarise the work and discuss its contribution towards addressing the motivating research questions.  

\section{Background and related work}
\label{sec:Background}

In contrast to the broader problem of hierarchical text classification, coarse-grained class labels are already available in the $FGSI$ task, which is defined as the refinement of existing coarse-grained labels in this work.
Recently, Mekala \textit{et al.}~\cite{Mekala2021} focused on this special case 
proposing the \textit{Coarse2Fine} ($C2F$) method.
$C2F$ relies on the literal occurrence of fine-grained labels as initial weak fine-grained labels, which are used together with the ground-truth coarse-grained ones to train a label-conditioned GPT2~\cite{AlecRadford2020} model that generates relevant documents for any given label.
This model is used to generate new documents for each fine-grained label, which are, in turn, used to train \textit{BERT} classifiers~\cite{Devlin2019}. 

Semantic indexing of biomedical literature in MEDLINE/PubMed has been done for several decades by human expert indexers of NLM. 
Since 2002 some automated topic suggestions are offered to human experts by the \textit{Medical Text Indexer} (\textit{MTI}) Tool\footnote{\url{https://lhncbc.nlm.nih.gov/ii/tools/MTI.html}}, and since 2012 a dedicated shared task has been organized for this purpose, in the context of the BioASQ challenge~\cite{tsatsaronis2015overview}. 
Despite the particular challenges of this task, such as the huge number of classes, the availability of millions of manually indexed articles and advanced DL architectures allowed the gradual development of machine-learning models that achieve satisfactory results~\cite{Nentidis2022,you2021bertmesh,Rae2021clef}.
As a result, NLM gradually moved to fully automated indexing for all MEDLINE articles\footnote{\url{https://www.nlm.nih.gov/pubs/techbull/nd21/nd21_medline_2022.html}}. 
However, indexing at the level of MeSH concepts still remains beyond reach, partly due to the lack of adequate data.    

Early work towards $FGSI$ of biomedical literature focused on enriching the descriptor-based labels with MeSH qualifiers and MeSH Supplementary Concept Records (SCRs)\footnote{\url{https://www.nlm.nih.gov/mesh/intro\_record\_types.html}}~\cite{Aronson2004}, which however do not address the coarseness of the descriptors that aggregate distinct concepts.
The importance of indexing the biomedical literature at the level of MeSH concepts was first noted in\cite{Darmoni2012}. 
More recently, we proposed the \textit{Beyond MeSH} method \cite{Nentidis2019_CBMS, nentidis2020beyond} for the automated refinement of MeSH topic annotations at the level of MeSH concepts. 
In that work, we introduced the \textit{concept-occurrence} ($CO$) heuristic for assigning weak labels for $FGSI$. This heuristic extends simpler dictionary-based ones
using the \textit{MetaMap} tool, which handles information extraction challenges such as inflected and multi-word terms, and concept disambiguation \cite{Aronson2010}. 
Experiments with common machine-learning algorithms on small datasets for refining two specific descriptors showed that the $CO$ heuristic is a strong baseline for these two use cases and only some trained classifiers managed to improve upon it.

Building upon this earlier work, in this paper we perform an extensive evaluation of the promising $CO$ heuristic in a large-scale scenario for several MeSH descriptors, while also assessing the performance of DL approaches for the first time in this task, revealing the particular challenges of the $FGSI$ task and proposing respective remedies.
In order to form the large-scale experimental scenario, we look retrospectively into the evolution of MeSH descriptors, focusing on MeSH concepts that over time become descriptors themselves. 

\section{Material and methods}
\label{sec:Methods}

\begin{figure}[!h]
\begin{center}
\includegraphics[width=1\textwidth]{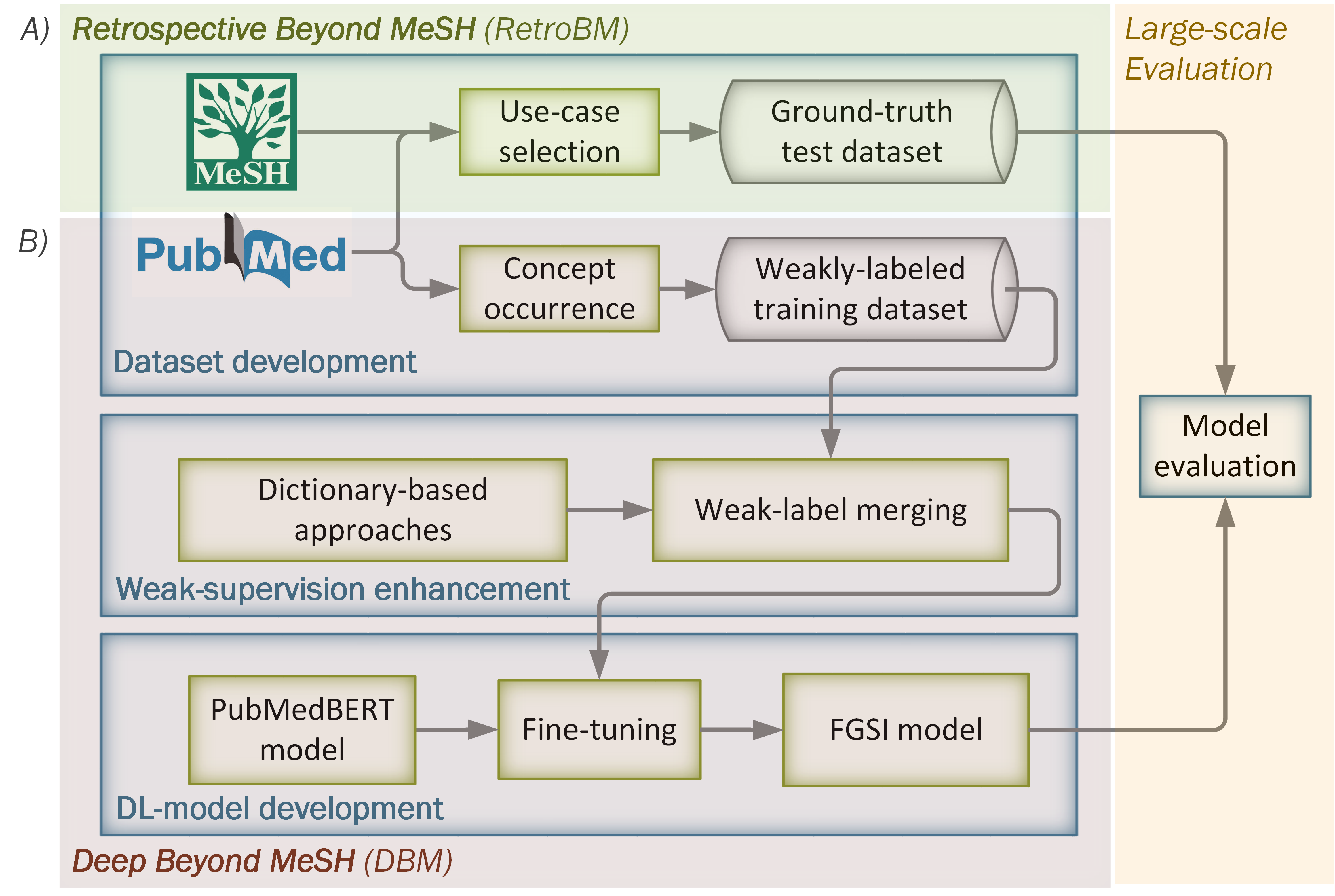}
\end{center}
\caption{A) The $RetroBM$ method for the development of large-scale ground-truth $FGSI$ datasets and B) the $DBM$ method for weakly-supervised $FGSI$ model development. 
The $RetroBM$ datasets are used for the large-scale evaluation of the $DBM$ models. }
\label{fig:DeepBeyondMeSH_block}       
\end{figure}
In this section, we introduce a) the \textit{Retrospective Beyond MeSH} ($RetroBM$) method for the development of ground-truth datasets for large-scale evaluation on Fine-Grained Semantic Indexing ($FGSI$), and b) the \textit{Deep Beyond MeSH} ($DBM$) method for refining coarse-grained MeSH labels into fine-grained ones, for which no ground-truth data are available.

As shown in Figure~\ref{fig:DeepBeyondMeSH_block}, part A, $RetroBM$ first selects some adequate use cases of MeSH concepts promoted to descriptors, by examining retrospectively the evolution of MeSH. Then it develops ground-truth $FGSI$ datasets for the task of refining these coarse-grained descriptors, based on the manual annotation already available in MEDLINE/PubMed. 

$DBM$, on the other hand (Figure~\ref{fig:DeepBeyondMeSH_block}, B), provides $FGSI$ for any subordinate MeSH concept, regardless of its promotion as a descriptor, without access to ground-truth labels. 
$DBM$ is structured into three parts: a) the development of weakly-labelled training datasets based on \textit{concept occurrence} ($CO$), b) the enhancement of weak supervision using dictionary-based approaches, and c) the development of $FGSI$ models, by fine-tuning pre-trained \textit{PubMedBERT} models. 

Finally, the ground-truth $RetroBM$ datasets are used both for the validation of the $DBM$ method and the evaluation of weakly-supervised $DBM$ models for the selected use cases. 
These methods are openly available in GitHub\footnote{\url{https://github.com/tasosnent/DBM}}. 

\subsection{Dataset development}
Although the proposed $DBM$ method develops weakly-labelled datasets to sidestep the need for ground-truth labels in model development, some ground-truth data are still needed for the validation and evaluation of such $DBM$ models.
Therefore, we also introduce $RetroBM$, an approach based on the evolution of MeSH indexing already done in MEDLINE/PubMed, to develop some ground-truth $FGSI$ datasets. 

\begin{figure}[!t]
\includegraphics[width=0.95\textwidth]{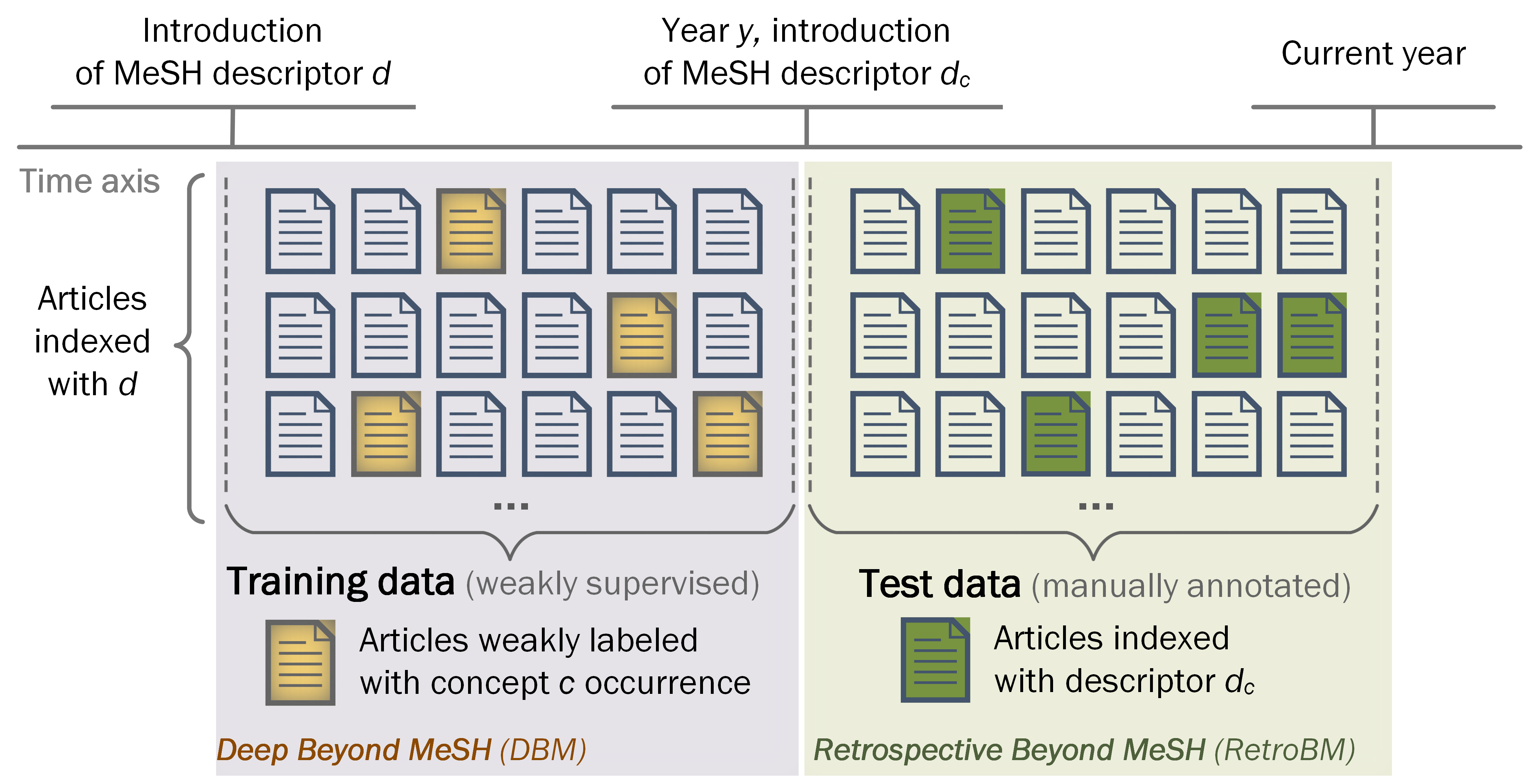}
\caption{
The retrospective evaluation scenario adopted in this work for the development of $FGSI$ datasets is based on the promotion of a MeSH concept \textit{c} into a descriptor $d_c$.}
\label{fig:retro_schema}       
\end{figure}

The two methods complement each other in a retrospective evaluation scenario based on the promotion of a MeSH concept \textit{c} to a dedicated descriptor $d_c$ in a year \textit{y} (Figure~\ref{fig:retro_schema}).
In this scenario, $RetroBM$ uses articles indexed with $d$ after the year \textit{y} to develop a ground-truth testset dataset, and $DBM$ uses articles indexed with $d$ prior to year \textit{y}, to develop a respective weakly-labelled training dataset\footnote{As articles indexed with \textit{d} we consider both articles directly annotated with the descriptor \textit{d} and articles annotated with any descriptor narrower to \textit{d} in the MeSH hierarchy.}. This is done for simplicity and alignment with a prospective scenario of use, where we develop a model on existing articles indexed with $d$ and use it in new ones. However, new articles could be weakly labelled and used by $DBM$ for model training as well.

\subsubsection{\textit{Retrospective Beyond MeSH} ($RetroBM$)}
\label{sssec:retro_data_scenario}

MEDLINE/PubMed provides annotated data for several parent-child pairs of MeSH descriptors. However, not all such pairs represent realistic use cases for evaluating the $FGSI$ task where existing coarse-grained annotations need to be refined into fine-grained ones.
For example, there are cases where the child descriptor is older than its parent, leading to a situation where annotations for the child are already available and no refinement of parent annotations is actually needed. For this reason, $RetroBM$ relies on the evolution of MeSH to identify suitable and realistic concept-promotion cases for $FGSI$ evaluation. 

\begin{figure}[!t]
\centerline{\includegraphics[width=0.87\textwidth]{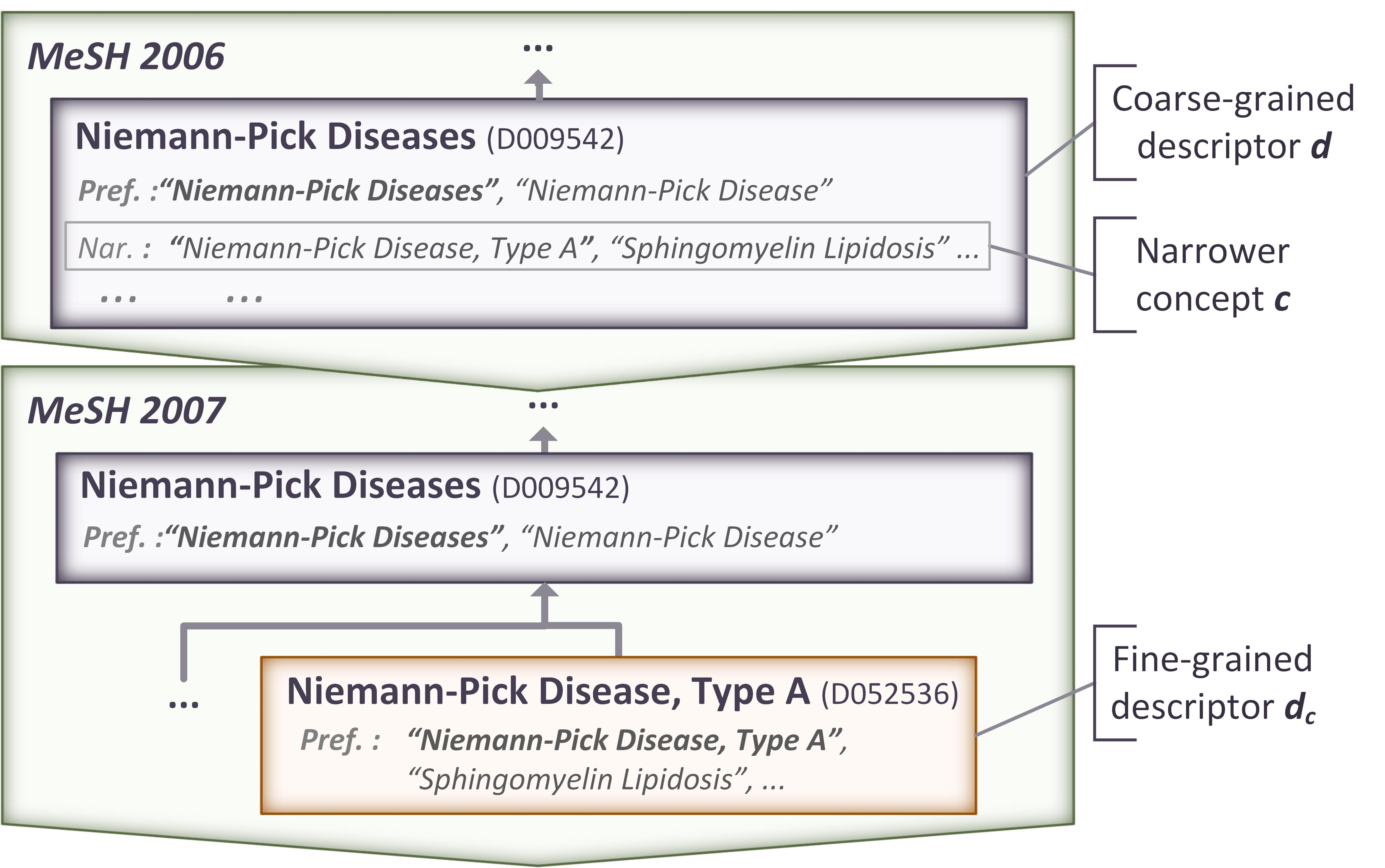}}
\caption{ The promotion of ``Niemann-Pick Disease, Type A'', from a subordinate concept of the descriptor ``Niemann-Pick Diseases'', to a new dedicated descriptor in 2007.  }\label{fig:MeSH_subdivision_Example}
\end{figure}

During the annual update of the MeSH thesaurus for a year \textit{y}, a subordinate concept \textit{c} of a coarse-grained descriptor \textit{d} that becomes increasingly important in the literature, can be promoted to a new MeSH descriptor \textit{d\textsubscript{c}}. 
In the example of Figure \ref{fig:MeSH_subdivision_Example}, ``Niemann-Pick Disease, Type A'' 
was a narrower subordinate concept (\textit{c}) of the ``Niemann-Pick Diseases'' descriptor (\textit{d}) until 2007 (\textit{y}), when it was promoted to a descriptor (\textit{d\textsubscript{c}}) itself. 
As a result, since 2007, articles relevant to ``Niemann-Pick Disease, Type A'' are indexed with a new dedicated descriptor (\textit{d\textsubscript{c}}).
In such cases, if \textit{c} is the only concept of the new descriptor \textit{d\textsubscript{c}}, indexing with \textit{d\textsubscript{c}} is equivalent to indexing with \textit{c}. 
Therefore, articles manually indexed after the promotion of \textit{c} into \textit{d\textsubscript{c}} have ground-truth labels for $FGSI$ with \textit{c} and can be used by $RetroBM$ to develop a testset, as illustrated in Figure~\ref{fig:retro_schema}. 

In particular, the task of $FGSI$ with \textit{c} is to identify for each article already indexed with \textit{d} (e.g.``Niemann-Pick Diseases''), whether it is about \textit{c} (e.g. ``Niemann-Pick Disease, Type A'') or not.
Therefore, we consider only articles indexed with \textit{d} and we refer to these articles as \textit{valid} articles for $FGSI$ with \textit{c}. 
A ground-truth dataset generated by $RetroBM$ consists of all these \textit{valid} articles annotated after year \textit{y}, where the ones indexed with \textit{d\textsubscript{c}} are the positive instances, and the remaining ones are the negative instances for $FGSI$ with \textit{c}.
Of course, indexing with \textit{d}, is not a sufficient condition for indexing with \textit{d\textsubscript{c}}, it is, however, a necessary condition, as \textit{d\textsubscript{c}} is a subcategory of \textit{d}. 
On the other hand, articles not indexed with \textit{d} (non \textit{valid} for \textit{c}) are beyond the scope of the $FGSI$ task, as they can't be relevant to its subcategory \textit{d\textsubscript{c}} either. 

Based on the above idea, $RetroBM$ looks retrospectively to identify several such concept-promotion events and develop corresponding datasets, introducing two criteria. 
First, $RetroBM$ relies on a typology that we introduced recently about the provenance of new MeSH descriptors~\cite{Nentidis2021}.
Out of the types introduced in this work, the one named \textit{subdivision}\footnote{We refer to cases with provenance code 1.2 as introduced in Nentidis \textit{et al.} \cite{Nentidis2021}. 
} represents subordinate concepts \textit{c} promoted to subcategories \textit{d\textsubscript{c}} of the coarse-gained descriptor \textit{d} that used to host them. 
This type satisfies the scenario of Figure~\ref{fig:MeSH_subdivision_Example} and was used by $RetroBM$ to identify adequate cases (\textit{criterion 1}). The presence of additional parents for \textit{d\textsubscript{c}} does not affect this scenario, as \textit{c} always belongs to a single descriptor \textit{d}. 
However, by considering other types of evolution, RetroBM can be extended to similar problems, such as indexing with emerging concepts, concepts with multiple parents, or even coarse-grained indexing with groups of concepts. 

Beyond provenance, each new descriptor \textit{d\textsubscript{c}} in $FGSI$ also needs to be fine-grained (\textit{criterion 2}), which breaks into the following: a) \textit{d\textsubscript{c}} should be dedicated to the corresponding promoted concept \textit{c}, not covering other subordinate concepts as well, ensuring that all articles annotated with \textit{d\textsubscript{c}} are actually relevant to \textit{c} (\textit{criterion 2.1}).
b) \textit{d\textsubscript{c}} should be a leaf in the MeSH hierarchy. Having narrower descriptors representing subcategories indicates that the descriptor covers several concepts, hence it is not fine-grained (\textit{criterion 2.2}).   

\subsubsection{Weakly-labelled training datasets}
The first step of the $DBM$ method uses articles annotated with any coarse-grained \textit{d} to develop weakly-labelled training data for $FGSI$ with any subordinate concept \textit{c}, without any need for manual \textit{d\textsubscript{c}} annotations.
In order to generate the weak labels $DBM$ uses the heuristic introduced in \textit{Beyond MeSH}~\cite{Nentidis2019_CBMS,nentidis2020beyond}, which relies on the occurrence of concept \textit{c} in the title or abstract of the article, as identified by the \textit{MetaMap} tool \cite{Aronson2010}. 
$DBM$ is applicable to any coarse-grained descriptor \textit{d\a'} having at least one narrower subordinate concept \textit{c\a'}, in order to predict labels for $FGSI$ with \textit{c\a'}, regardless of whether \textit{c\a'} has been promoted to a descriptor or not. 
In MeSH 2020, for example, there are about 7,000 such narrower MeSH concepts with sufficient weak supervision for applying $DBM$\footnote{For this estimation, we consider any narrower concept \textit{c\a'} of a descriptor \textit{d\a'} with at least 10 articles having a) \textit{d\a'} annotation, and b) \textit{c\a'} occurrence.}.
However, ground-truth \textit{c\a'} labels will be needed to verify the accuracy of this annotation process.
For this reason, in this work, we employ $DBM$ for use cases selected by $RetroBM$ as illustrated in the evaluation scenario of Figure~\ref{fig:retro_schema}.  

\subsubsection{Multi-label setup}
The $FGSI$ task is inherently multi-label as a single article can be relevant to several distinct coarse-grained and fine-grained labels. 
In addition, in $FGSI$ we would like any method to refine a single descriptor (e.g. ``Epilepsy'') into any of its sibling subordinate concepts (e.g. ``Audiogenic Epilepsy'', ``Tactile Reflex Epilepsy'' etc.), which are not usually mutually exclusive.
In this work, aligned with the retrospective evaluation scenario, we organize the $FGSI$ datasets developed by $RetroBM$ and $DBM$ based on the promotion year \textit{y}, gathering all the \textit{valid} articles for any of the use cases corresponding in a given year \textit{y} together and assigning multi-label annotations accordingly.

\subsection{Weak-supervision enhancement}
\label{ssec:WS_study}
The second part of the $DBM$ method is the enhancement of the original weak supervision provided by $CO$.
In this section, we investigate a range of dictionary-based variants and whether their combination could enhance the quality of the weak labels. The results of this study on a few validation cases, presented in Subsection~\ref{ssec:val_experiments}, allow the selection of the strategy to be adopted in the second step of the $DBM$ method for the enhancement of weak supervision.
In particular, adopting the terminology introduced in data programming~\cite{Ratner2016}, we refer to these dictionary-based variants and $CO$ as labeling functions (LFs) for the $FGSI$ task.

Each dictionary-based labeling function (LF) assigns a label for concept \textit{c} to an article if any dictionary element associated with \textit{c} literally occurs in the title or abstract of the article\footnote{An LF with n dictionary elements ($e_i$) is implemented with the query: $e_1$ or $e_2$ ... or $e_n$.}. 
In the \textit{name exact} LF only the name of \textit{c} is matched (e.g.``Niemann-Pick Disease, Type A''), and in the \textit{synonyms exact} the synonyms of it (e.g.``Classical Niemann-Pick Disease'', etc).
Then, six variants of these two LFs are introduced\footnote{For applying the latter six LFs, the title and abstract of the article are also lower-cased.}, 
in order to improve recall (see Table~\ref{table:Label_functions}). 
In particular, the dictionary elements are first lower-cased (\textit{name lowercase}, \textit{synonyms lowercase}), then the punctuation marks are removed (\textit{name no punct}, \textit{synonyms no punct}), and finally, phrases are split into single-token elements (\textit{name tokens}, \textit{synonyms tokens}).

These dictionary-based LFs rely on the same basic information that is used by $CO$, that is the terms of \textit{c} and the article text. Still, they may  result in different weak labels. 
This is because $CO$, calculated by \textit{MetaMap}, is more elaborate than exact term matching, considering additional information such as part-of-speech tagging and word-sense disambiguation based on context~\cite{Aronson2010}. 
Therefore, combining these LFs with $CO$ may bring improvements.
The correlation between these LFs, on the other hand, may harm the accuracy of the ensemble.
For this reason, the optimal LF subset is sought for adoption in $DBM$, by considering all the combinations of two or more of these LFs in the validation experiments (Subsection~\ref{ssec:val_experiments}). 

\begin{table}[!t]
\caption{The dictionary elements of each LF for the label ``Niemann-Pick Disease, Type A''.}
\begin{center}
\small
\begin{tabular}{L{0.28\linewidth}L{0.1\linewidth}L{0.5\linewidth}}
Labeling function (LF)  & Abbr. & Dictionary elements                        \\
\hline
\textit{name exact}        & $NE$ &``Niemann-Pick Disease, Type A''                        \\
\textit{synonyms exact}    & $SE$ &``Classical Niemann-Pick Disease'', ...\\
\textit{name lowercase}    & $NL$ &``niemann-pick disease, type a''                                  \\
\textit{synonyms lowercase} & $SL$ &``classical niemann-pick disease'', ...\\
\textit{name no punct}     & $NNP$ &``niemann pick disease type a''                                  \\
\textit{synonyms no punct} & $SNP$ &``classical niemann pick disease'', ...\\
\textit{name tokens}       & $NT$ &``niemann'', ``pick'', ``disease'',  ``type'', ``a'' \\
\textit{synonyms tokens}    & $ST$ &``classical'', ``niemann'', ``pick'', ``disease'' ...
\end{tabular}
\label{table:Label_functions}
\end{center}
\end{table}

For combining the LFs into an ensemble we examine two voting classifiers based on \textit{majority voting} ($MV$) and \textit{at-least-one} voting ($ALO$)~\cite{kambhatla2006minority}, as well as a state-of-the-art probabilistic \textit{label modeling} ($LM$) approach for combining noisy heuristics, introduced in the context of Snorkel~\cite{Ratner2017}. 
The $MV$ schema assigns a label to an article if most of the LFs (voters) assign this label to this article, favoring the precision of the ensemble, as labels predicted by only a few LFs are disregarded as untrustworthy.
In the $ALO$ schema, on the other hand, a label is assigned to an article if any of the LFs assign this label to this article. This is a special case of the \textit{at-least-N} schema, proposed by Kambhatla~\cite{kambhatla2006minority}, where $N=1$.
In this approach, even if a single labeling function predicts a label, this will be included in the prediction of the ensemble, favoring its recall.
Finally, the Snorkel $LM$ approach considers the true label for an article as a latent variable in a probabilistic generative model, which is estimated based on a weighted combination of the LFs~\cite{Ratner2016, Ratner2017}. Eventually, a label is assigned to an article if the estimated probability of the label is greater than 0.5.
     

\subsection{Model development}
\label{ssec:model_dev}
The final part of the $DBM$ method is the development of weakly-supervised Deep Learning (DL) models for the $FGSI$ task. 
In this direction, motivated by experiments on other tasks suggesting that deep pre-trained models can have an inherent resilience to label noise~\cite{Tanzer2021, Hendrycks2019}, we investigate the adequacy of a state-of-the-art approach in text classification, that is fine-tuning a pre-trained \textit{BERT}-based model, in our case a \textit{PubMedBERT} model~\cite{Gu2022}, with one additional task-specific output layer~\cite{Devlin2019}.  
In particular, we experimented on a few validation cases, in order to identify the particular challenges of the $FGSI$ task and adopt remedies to tackle them. These include preliminary experiments for basic design choices discussed in this section, as well as the validation experiments presented in Subsection-\ref{ssec:val_experiments}.

\begin{figure}[!t] \includegraphics[width=0.95\textwidth]{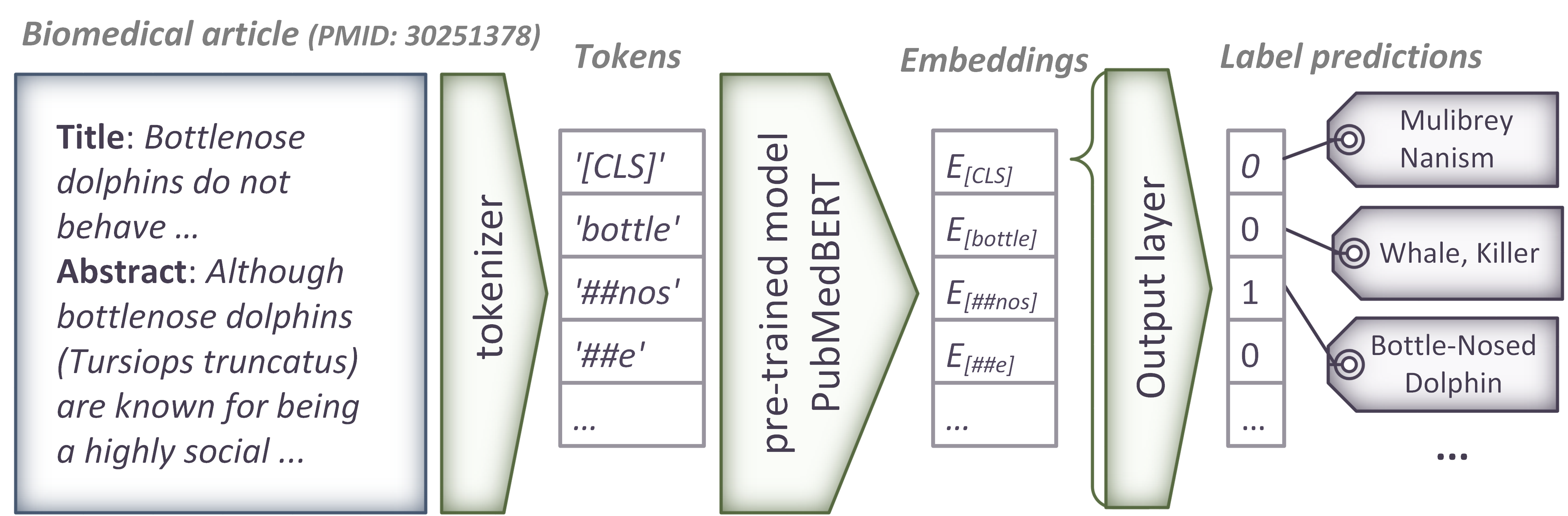}
\caption{A multi-label model is developed by $DBM$ for all the labels of a single year.}
\label{fig:BERT_ML}       
\end{figure}

In $DBM$, we opt for a multi-label scenario, where a distinct multi-label model is developed for each year \textit{y}, as depicted in Figure~\ref{fig:BERT_ML}\footnote{A common model for many years is possible but would pose experimental complications as a single article could end up in the test set of one use case and the training set of another.}.   
In this architecture, the title and abstract of each article are concatenated, tokenized, and provided as input to a \textit{PubMedBERT} model. The embedding of the classification token $CLS$ is then provided as input in a fully-connected output layer that produces a vector of size equal to the number of labels as in~\cite{DBLPSilvaG21}. 
This way, knowledge from different labels is integrated into shared deep representations for all the use cases of a single year. 
This can be useful, particularly for sibling concepts of a single coarse-grained descriptor, such as ``Audiogenic Epilepsy'' and ``Tactile Reflex Epilepsy'' in the example of Figure~\ref{fig:MeSH_Example}.    
Preliminary experiments confirmed that multi-label models are at least as good as single-label models in a One-Versus-Rest approach. Therefore, the additional cost of developing and storing multiple single-label models was not justified. 

Regarding the specific \textit{BERT}-based model to fine-tune, we opted for \textit{PubMedBERT}~\cite{Gu2022}, a domain-specific variant of \textit{BERT} adopting a domain-specific vocabulary to better capture the semantics of the text. 
Although differences in the preliminary results were small, they confirmed the good performance of \textit{PubMedBERT} and it was thus chosen as the basis for all $DBM$ models.    
The preliminary experiments also indicated that the number of training epochs can affect the performance of the model. 
For this purpose, we split the weakly-labelled training data into a \textit{training part} and a \textit{validation part} (90-10\%) in order to train only on the first and check the loss in the unseen \textit{validation part}, identifying in which epoch this loss is minimised (\textit{best\_ep}).

For a higher generalization level, we also adopted an early stopping strategy that concludes the fine-tuning process once the loss on the \textit{validation part} starts increasing~\cite{prechelt1998early}.
In this case, the $best\_ep$ model is always that of the penultimate epoch of the early stopped fine-tuning.  
Furthermore, we also considered the $prev\_ep$ and $next\_ep$ models that correspond to the epochs just before or after $best\_ep$. The motivation was that the ``imperfection" of epochs close to the best-performing one may be due to a higher level of generalization. 

A major challenge in training $FGSI$ models is the imbalance of the labels in two ways: a) The number of positive instances for each label varies among use cases, favoring more popular fine-grained labels, and b) there are many more negative instances than positive ones. In extreme cases, this abundance of negative instances can prevent the models from learning to recognize positive ones, as their overall effect on the loss is minimal. 

For imbalance among labels, we considered extensions of the well-established binary cross entropy (BCE) loss~\cite{Bengio2013} that can handle imbalanced training data for multi-label text classification~\cite{Huang2021}. 
In particular, the rebalanced focal loss (R-BCE-FL) is an extension of BCE loss, first introduced to handle label imbalance in multi-label image classification tasks~\cite{Wu2020} and recently extended to multi-label text classification tasks~\cite{Huang2021}.    
R-BCE-FL modifies BCE in two ways. First, the instances are rebalanced, by oversampling based on the overall frequency of corresponding labels. Second, a focusing parameter increases the weight of more hard-to-classify instances, based on their predicted probability~\cite{Huang2021}. 

For the imbalance between positive and negative instances, previous work has shown that under-sampling can help address this problem~\cite{nentidis2020beyond}. 
Therefore, we experimented with an under-sampling procedure removing negative instances to achieve an ideal negative-to-positive ratio ($balance\_n$) for each label.
This ideal ratio is not known beforehand, but a reasonable estimate can be provided through validation. 
Removing random negative instances to enforce the ideal negative-to-positive ratio for a label strictly, may deplete the positive or \textit{valid} negative instances for another. 
For this reason, the undersampling process of $DBM$ does not remove instances that are either positive for any of the labels or are \textit{valid} for labels with a low negative-to-positive ratio.

Randomness is introduced in several steps of the model development process~\cite{Dodge2020}, notably in the splitting of training and validation parts, in the under-sampling of negative instances, and in stochastic optimization~\cite{Kingma2014}. 
The preliminary experiments showed that starting with different random seeds can produce different results. For this reason, each experiment was repeated with six different seeds, and the predictions were aggregated with a majority voting approach, which led to a more stable estimate of model performance.

\subsection{Competing machine learning approaches}
\label{ssec:baselines}

Three machine-learning approaches have been included in our experiments for comparison, based on the most recent related work (Section~\ref{sec:Background}). 
First, \textit{Beyond MeSH}~\cite{nentidis2020beyond} was considered, as a prior work on the $FGSI$ task of refining MeSH topic annotations at the level of MeSH concepts. 
In this direction, a logistic regression ($LR$) model was developed for each fine-grained label considering both lexical and semantic features.
For a fair comparison with $DBM$, we perform the same under-sampling of negative instances, which is repeated for six seeds combining the results with majority voting.
The level of regularization is selected via grid search on a \textit{validation part} (10\%) of the weakly-labelled data, considering a range of values from 0.01 to one billion\footnote{Namely these values were considered: 0.01, 0.1, 1, 10, 100, 1K, 10K, 100K, 1M, 1B.}.
A reasonable number of features to be selected was chosen based on the validation experiments presented in~\ref{ssec:val_experiments}. 

Second, \textit{Coarse2Fine} ($C2F$)~\cite{Mekala2021} was also considered, as a state-of-the-art work on the broader task of refining existing coarse-grained labels into fine-grained classes. 
Although the $C2F$ originally relies on the literal occurrence of label names, for a fair comparison against $DBM$, we provided $C2F$ with the same weak supervision.
In addition, $C2F$ handles multi-class problems, where each document belongs to exactly one of the fine-grained sub-categories of a coarse-grained label \textit{d}. 
In order to handle the ``unlabelled'' examples in $FGSI$, that belong to \textit{d} but not to any of its sub-categories, we introduce an additional subcategory (\textit{d\_rest}) for each \textit{d}. 
The predictions for \textit{d\_rest}, however, are disregarded when estimating the performance of the $C2F$ models, which is based on the actual subcategories, as done for all methods.  
As $C2F$ adds generated instances to handle imbalance, we did not undersample the data.
In addition, we did not experiment with different random seeds or parameters relying on the original implementation\footnote{https://github.com/dheeraj7596/C2F}. 

Finally, in order to investigate the impact of generating instances as per $C2F$, we also developed $DBM^{C2F}$ models using the implementation of $DBM$ with the documents generated by $C2F$ as weak supervision.

\subsection{Evaluation measures}
As both precision ($P$) and recall ($R$) are important for $FGSI$, we adopt the F1-measure ($F1$) as an overall performance measure, considering both macro-averaging ($maF1$, $maP$, $maR$) and micro-averaging ($miF1$, $miP$, $miR$) across different use cases~\cite{Tsoumakas2009}.
In the multi-label $FGSI$ datasets articles that are not \textit{valid} for a specific fine-grained label \textit{c} are not of interest and are disregarded in the evaluation of that label. 
We refer to this process as \textit{validity filtering} and employ it in the evaluation of all the methods. 


\section{Experiments and results}
\label{sec:Experiments}

$RetroBM$ was applied to fourteen consecutive years (2006-2019) of MeSH evolution for developing respective ground-truth datasets (\textit{test\textsubscript{y}}). 
Then, $DBM$ developed respective weakly-labelled training datasets (\textit{WS\_dev\textsubscript{y}}) for $FGSI$-model development (Subsection~\ref{ssec:retro_data_2005-2020}).
Out of these datasets, we used the ones for labels introduced in 2006 (\textit{2006 datasets}) for ``validation experiments'', in order to make design choices about the architecture and configurations of the $DBM$ method (Subsection~\ref{ssec:val_experiments}). 
Finally, the datasets for unseen fine-grained labels introduced in the subsequent years (\textit{2007-2019 datasets}) were used in the ``evaluation experiments'' (Subsection~\ref{ssec:eval_experiments}), which aimed to confirm the choices made on the \textit{2006 datasets}.

\subsection{Retrospective datasets}
\label{ssec:retro_data_2005-2020}

In total, 88 new fine-grained descriptors were found to match the criteria of $RetroBM$, which was employed to retrospectively process MEDLINE data from 2006 to 2019\footnote{The full list of descriptors per year is available in the supplementary file S1. 
}.
In particular, the 2020 version of the MEDLINE/PubMed Annual Baseline Repository was used to retrieve documents and MeSH annotations. In addition, SemMedDBv4.0 was used by $DBM$ for $CO$ and use cases without enough data for training and evaluation were excluded from the experiments\footnote{We kept use cases with at least 10 articles with a) $c$ occurrence in \textit{WS\_dev\textsubscript{y}}, and b) $d_c$ annotations in \textit{test\textsubscript{y}}. A too-general use case with more than 1.5M articles for a single \textit{d} was also excluded for practical reasons.}.

In particular, $RetroBM$ identified 18 descriptors introduced in 2006 developing the \textit{test\textsubscript{2006}} ground-truth dataset and $DBM$ developed the respective weakly-labelled dataset \textit{WS\_dev\textsubscript{2006}}. 
These two datasets (\textit{2006 datasets}) consisted of more than 400 thousand articles each and were used for validation experiments\footnote{\textit{WS\_dev\textsubscript{2006}} 446,377 articles, \textit{test\textsubscript{2006}} 468,271 articles.}. 
The remaining 70 descriptors, introduced after 2006, were used for the evaluation experiments (\textit{2007-2019 datasets}). The total size of the thirteen weakly-labelled datasets (\textit{WS\_dev\textsubscript{2007-2019}}) and the respective ground-truth ones (\textit{test\textsubscript{2007-2019}}), exceeds 2M and 1M articles respectively\footnote{\textit{WS\_dev\textsubscript{2007-2019}} 2,068,965 articles, \textit{test\textsubscript{2007-2019}} 1,135,657 articles.}. 

In these datasets, the positive instances are really scarce highlighting the importance of balancing approaches adopted in $DBM$.
In \textit{test\textsubscript{2006}}, in particular, the positive instances for any of the 18 labels are almost 6\% of all articles in \textit{test\textsubscript{2006}}.
In addition, about 3\% and 5\% of the articles in  \textit{WS\_dev\textsubscript{2006}} and \textit{test\textsubscript{2006}} are positive based on the weak labels. 
These frequencies are similar for the 70 labels of \textit{2007-2019 datasets} as well, considering both the weak and the ground-truth labels\footnote{Detailed statistics for each dataset are available in the supplementary file S1.}. This suggests that observations on older labels can be useful for labels introduced later, confirming the rationality of a retrospective scenario. 

\subsection{Validation experiments on \textit{2006 datasets}}
\label{ssec:val_experiments}

he first aim of the validation experiments was to confirm the strength of \textit{concept occurrence} ($CO$) as the weak-supervision source of $DBM$.
Table~\ref{table:LF_results} presents LF performance for the $FGSI$ task on \textit{test\textsubscript{2006}}.
$CO$ was clearly the strongest LF, which confirms previous observations on smaller-scale experiments~\cite{nentidis2020beyond}.
In addition, we observe that most LFs achieved better precision than recall, except those based on tokenization (\textit{NT}, \textit{ST})\footnote{
Some tokens in these LFs can be too-general such as ``disease'', ``type'', and ``a'' in Table~\ref{table:Label_functions}.}.
The average correlation between $CO$ and each other LF was 0.48 with $SL$, $NNP$, and $SNP$ being the most correlated (0.68) and $ST$ the least correlated (0.21) with $CO$. Similarly, the average correlation between the dictionary-based LFs themselves was 0.35, with [$SL$, $SNP$] being the most correlated pair (0.95) and [$SL$, $ST$] being the least correlated one (0.06)\footnote{The full correlation matrix for all LFs is available in the supplementary file S2.}.

\begin{table}[!t]
\caption{$FGSI$ results of each labeling function on the $test_{2006}$ ground-truth data. 
$var$ stands for the variance of $maF1$ across the 18 labels.}
\begin{center}
\small 
\begin{tabular}{lllllll}
Labeling function (LF)   & $maP$ & $maR$    & $maF1$\footnotesize{ $\pm var$}	& $miP$   & $miR$   & $miF1$\\
\hline
\textit{concept occurrence} (\textit{CO})	& 0.715& 0.618& \textbf{0.634}\footnotesize{  $\pm0.06$}&0.737&0.531&\textbf{0.618}	\\
\textit{name exact} (\textit{NE})     		& 0.707& 0.082& 0.112\footnotesize{ $\pm0.04$}&0.901&0.027&0.052	\\
\textit{synonyms exact} (\textit{SE})   	& 0.656& 0.212& 0.251\footnotesize{ $\pm0.12$}&0.889&0.046&0.088	\\
\textit{name lowercase} (\textit{NL})     	& 0.649& 0.441& 0.446\footnotesize{ $\pm0.12$}&0.589&0.203&0.302	\\
\textit{synonyms lowercase} (\textit{SL}) 	& 0.620& 0.365& 0.408\footnotesize{ $\pm0.11$}&0.709&0.251&0.371	\\
\textit{name no punct} (\textit{NNP})   	& 0.574& 0.440& 0.442\footnotesize{ $\pm0.13$}&0.583&0.196&0.293	\\
\textit{synonyms no punct} (\textit{SNP})	& 0.597& 0.349& 0.385\footnotesize{ $\pm0.11$}&0.690&0.206&0.318	\\
\textit{name tokens} (\textit{NT}) 			& 0.322& 0.838& 0.377\footnotesize{ $\pm0.07$}&0.157&0.807&0.263	\\
\textit{synonyms tokens} (\textit{ST})		& 0.274& 0.874& 0.338\footnotesize{ $\pm0.07$}&0.081&0.883&0.149	\\
\end{tabular} 
\end{center}
\label{table:LF_results}
\end{table}

The second aim of these experiments was to confirm whether $CO$ can be enhanced when ensembled with other LFs, choosing a) which LF subset and b) which ensemble approach to adopt in $DBM$ for weak-supervision enhancement. 
In this direction, for each LF subset, the three ensemble approaches were assessed for the $FGSI$ task on $test_{2006}$, namely \textit{majority voting} ($MV$), \textit{at-least-one} ($ALO$), and \textit{Snorkel label model} ($LM$).
The results confirmed that several LF subsets led to performance improvements upon $CO$ alone\footnote{The full results for all LFs subsets are available in the supplementary file S3.}.

Overall, two LF subsets\footnote{[$CO$, $NL$, $SL$] and [$CO$, $NL$, $SL$, $NNP$].} achieved the top $maF1$ (0.689) and $miF1$ (0.646) scores, under the $ALO$ or the $LM$ ensemble, outperforming $CO$ by 5.5pp and 2.8pp respectively. 
In addition, six more LF subsets led to these top scores under the $ALO$ approach only. 
Interestingly, $CO$, $NL$, and $SL$ are present in all top-performing LF subsets, regardless of the ensemble approach, and the combination of only these three LFs was sufficient to achieve top performance with both $ALO$ and $LM$.
As a result, the subset [$CO$, $NL$, $SL$] was adopted for weak supervision enhancement in $DBM$.

The fact that $ALO$ achieved top results with more LF subsets than $LM$\footnote{8 top-performance subsets with $ALO$ against 2 with $LM$.}, suggests that $ALO$ was more robust to the choice of LF subset in this setup.
A reason for this could be the high precision of most LFs and the fact that none of the top-performing subsets included the low-precision LFs ($NT$, $ST$), as $ALO$ is known to be good at combining precise voters by improving the total recall~\cite{kambhatla2006minority}. 
In particular, $ALO$ assigns a label to an article if at least one labeling function does, not affected by potential negative votes of more cautious LFs.
As a result, the $ALO$ ensemble of [$CO$, $NL$, $SL$] ($ALO_3$) was adopted as the enhanced weak supervision schema in $DBM$. 

Another aim of the validation experiments on \textit{2006 datasets} was to make specific choices finalizing the model-development part of $DBM$. In particular, we need to a) estimate a reasonable value for the negative-to-positive ratio ($balance\_n$), and b) choose the model of which epoch to use for $DBM$ predictions, that is, whether to use the model of the $best\_ep$ that minimized the loss on the \textit{validation part} of \textit{WS\_dev\textsubscript{y}}, or the epoch just before ($prev\_ep$) or just after $best\_ep$ ($next\_ep$).

In this direction, different $DBM$ models were developed on $WS\_dev_{2006}$ and assessed for their $FGSI$ performance on $test_{2006}$.
In particular, experiments with different $balance\_n$ values ranging from 1 to 100, led to the choice of 10-to-1 as a reasonable negative-to-positive ratio, which was adopted for all remaining experiments.
In addition, the $FGSI$ results of $DBM$ experiments with the three alternatives for the training epoch (Table~\ref{table:val_results}), suggest that the model corresponding to $next\_ep$ ($DBM_n$) led to the best results for both metrics. Therefore, the model of $next\_ep$ was adopted as the final one in $DBM$.

\begin{table}[!t]
\caption{$FGSI$ validation results on $test_{2006}$ data. 
WS stands for the source of weak supervision for trained models.
A subscript in $LR$ indicates the number of features selected.
A subscript in $DBM$ indicates the training epoch, with \textit{p}, \textit{b}, and \textit{n} corresponding to \textit{prev\_ep}, \textit{best\_ep}, and \textit{next\_ep} respectively.
$var$ stands for the variance of $maF1$ across the 18 labels.
}
\begin{center}
\small 
\begin{tabular}{llllllll}
LF/model   				& WS &  $maP$& $maR$& $maF1$\footnotesize{ $\pm var$}	& $miP$   & $miR$   & $miF1$\\
\hline	                     
$CO$ 					& - & 0.715& 0.618& 0.634\footnotesize{ $\pm0.06$} &0.737 &0.531 &0.618\\
$LR_{5}$    			& $CO$ & 0.427& 0.822& 0.474\footnotesize{ $\pm0.10$}&0.295 &0.655 &0.406\\
$LR_{10}$   			& $CO$ & 0.358& 0.807& 0.396\footnotesize{ $\pm0.09$}&0.102 &0.629 &0.176\\
$LR_{100}$  			& $CO$ & 0.275& 0.809& 0.318\footnotesize{ $\pm0.06$}&0.085 &0.620 &0.149\\
$LR_{1000}$ 			& $CO$ & 0.279& 0.798& 0.310\footnotesize{ $\pm0.07$}&0.134 &0.644 &0.221\\
$ALO_3$ 				& - & 0.717& 0.698& 0.689\footnotesize{ $\pm0.03$} &0.713 &0.591 &0.646\\
$DBM_{p}$  				& $ALO_3$ & 0.681& 0.744& 0.700\footnotesize{ $\pm0.03$} &0.689 &0.624 &0.655\\
$DBM_{b}$ 				& $ALO_3$ & 0.688& 0.738& \textbf{0.702}\footnotesize{ $\pm0.03$}	&0.690 &0.626 &0.656\\
$DBM_{n}$  				& $ALO_3$ & 0.689& 0.741& \textbf{0.702}\footnotesize{ $\pm0.03$}	&0.695 &0.623 &\textbf{0.657}\\
\end{tabular} 
\end{center}
\label{table:val_results}
\end{table}

Finally, validation experiments with logistic regression ($LR$) models on the same $FGSI$ task (\textit{2006 datasets}), also presented in Table~\ref{table:val_results}, allowed the selection of the number of features to be considered in the evaluation experiments. In particular, $LR$ models were trained on $WS\_dev_{2006}$ considering different numbers of features for feature selection with the F-ANOVA statistic, namely 5, 10, 100, and 1000. 
The results reveal that no $LR$ model managed to outperform $CO$ overall.
In particular, the best $LR$ model, namely $LR_5$ with five features, managed to outperform $CO$ in $F1$ for about 40\% of the 18 labels on $test_{2006}$\footnote{Detailed validation results per descriptor are available in the supplementary file S4.}.

\subsection{Evaluation on new labels introduced after 2006}
\label{ssec:eval_experiments}

In this last set of experiments, we assessed $DBM$ and the competing methods on $FGSI$ with the 70 unseen labels of \textit{2007-2019 datasets}. 
The enhancement of $CO$ supervision into $ALO_3$ is a key part of $DBM$. However, in order to investigate $ALO_3$ importance we also trained $DBM$ models directly on $CO$ alone ($coDBM$).
For the same reason, we trained the competing models on both the original ($CO$) and the enhanced ($ALO_3$) supervision.
The results of these experiments on the $FGSI$ task are presented in Table~\ref{table:eval_results}\footnote{Detailed evaluation results per descriptor are available in the supplementary file S5.}.

\begin{table}[!t]
\caption{$FGSI$ evaluation results on $test_{2007-2019}$ data. 
WS stands for the source of weak supervision for trained models.
A subscript in $LR$ models indicates the number of features selected.
A subscript in $DBM$ models indicates the training epoch considered (\textit{n} for \textit{next\_ep}).
$var$ stands for the variance of $maF1$ across the 70 labels.}
\begin{center}
\small 
\begin{tabular}{llllllll}
LF/model   			& WS & $maP$ & $maR$    & $maF1$\footnotesize{ $\pm var$}	& $miP$   & $miR$   & $miF1$\\
\hline                 
					   
$CO$				& - & 0.766& 0.590& 0.626\footnotesize{ $\pm0.06$}  			& 0.848 &0.554	&0.670\\
$NL$				& - & 0.647& 0.369& 0.408\footnotesize{ $\pm0.12$}  			& 0.880 &0.217	&0.348\\
$SL$				& - & 0.567& 0.253& 0.287\footnotesize{ $\pm0.09$}  			& 0.809 &0.125	&0.217\\
$LR_{5}$			& $CO$ & 0.420& 0.807& 0.458\footnotesize{ $\pm0.09$}  			& 0.204	&0.783	&0.324 \\
$C2F$				& $CO$ & 0.206& 0.946& 0.276\footnotesize{ $\pm0.09$}  			& 0.074 &0.981	&0.138\\
$coDBM_n$				& $CO$ & 0.718& 0.647& 0.641\footnotesize{ $\pm0.06$}  			& 0.826 &0.573	&0.677\\
$coDBM^{C2F}_{n}$  	& $CO$ & 0.712& 0.626& 0.625\footnotesize{ $\pm0.06$} &0.831 &0.568 &0.675\\
$MV_3$  			& - & 0.782& 0.481& 0.542\footnotesize{ $\pm0.08$}  			& 0.873 &0.281	&0.462\\
$LM_3$  			& - & 0.756& 0.652& 0.661\footnotesize{ $\pm0.06$}  			& 0.842    &0.569	&0.679\\
$ALO_3$ 			& - & 0.764& 0.668& \textbf{0.677}\footnotesize{ $\pm0.05$}  	& 0.839 &0.583	&0.688\\
$LR_5$ 			& $ALO_3$ & 0.403& 0.825& 0.454\footnotesize{ $\pm0.09$} & 0.196	&0.797	&0.315\\
$C2F$				& $ALO_3$ & 0.203& 0.959& 0.272\footnotesize{ $\pm0.09$}  			& 0.076 &0.983	&0.141\\
$DBM_n$ 			& $ALO_3$ & 0.725& 0.695& 0.672\footnotesize{ $\pm0.04$}  			& 0.839 &0.621	&\textbf{0.714}\\
$DBM^{C2F}_{n}$  	& $ALO_3$ & 0.726& 0.679& 0.656\footnotesize{ $\pm0.05$} &0.803 &0.588 &0.679
\end{tabular}
\end{center}
\label{table:eval_results}
\end{table}

These results confirm that $CO$ is a strong heuristic for $FGSI$, outperforming again the dictionary-based heuristics.
In addition, the weak-supervision enhancement schema ($ALO_3$) achieved an overall $maF1$ about 5pp higher than $CO$ alone, which is statistically significant\footnote{Wilcoxon signed-rank test, $H_1: F1\:CO < F1\:ALO_3$, p-value = 0.000001.}. 
The performance of the \textit{Snorkel Label model} of the same LF subset ($LM_3$) was not significantly lower than that of $ALO_3$, as the two approaches achieved practically the same $F1$ score for most of the 70 labels. However, $ALO_3$ achieved $F1$ performance equal to or better than that of $LM_3$ for all 70 labels.

Regarding trained models, the results show that logistic regression ($LR$) models did not manage to outperform the supervision provided to them overall, neither $CO$ nor $ALO_3$.
The overall $maF1$ performance of the $C2F$ models was also low, achieving top recall but low precision. 
In particular, $C2F$ trained on $CO$ did better than $CO$ itself for only 13 out of the 70 labels.
The low performance of a state-of-the-art method like $C2F$ is an indication that the $FGSI$ task and datasets have particular challenges, such as the high abundance of ``unlabelled'' documents that do not belong to any of the labels of interest. 

On the other hand, the $coDBM_n$ model developed using the weak supervision of $CO$ alone led to a limited overall performance gain of about 1.5pp in terms of $maF1$ over $CO$ itself.
The performance of the $DBM_n$ model, using the \textit{enhanced supervision} ($ALO_3$), was the best trained model, achieving about 4.6pp improvement over $CO$ in $maF1$, which is statistically significant\footnote{Wilcoxon signed-rank test, $H_1: F1\:CO < F1\:DBM_n$, p-value = 0.0065.}. 
However, even $DBM_n$ achieves only comparable performance to $ALO_3$ itself. In fact, $ALO_3$ alone has a slightly higher $maF1$, without statistical significance\footnote{Wilcoxon signed-rank test, $H_1: F1\:ALO_3 \neq F1\:DBM_n$, p-value = 0.088.}. 
Overall, the $DBM$ method, starting from $CO$ as a strong heuristic, managed to enhance it further and achieved the refinement of several coarse-grained labels to the level of 70 unseen fine-grained concepts, without access to ground-truth labels.   
These results suggest that some DL models adopting the design choices of $DBM$ can indeed improve upon the weak supervision that they rely on. 

\section{Discussion and conclusions}
\label{sec:Conclusion}
In this work, we investigate the task of fine-grained semantic indexing ($FGSI$), without ground-truth training data, and the development of weakly-supervised deep-learning (DL) models for it. Additionally, we propose an automated approach to generating large-scale ground-truth datasets for $FGSI$, based on MEDLINE articles. 
Hence, the main contributions of the paper are: 
\begin{itemize}
    \item The confirmation of \textit{concept occurrence} ($CO$) as a strong heuristic for $FGSI$ at a large-scale.
    \item A new weakly-supervised method, \textit{Deep Beyond MeSH} ($DBM$), for the $FGSI$ task. This method introduces a novel schema for enhancing the original weak supervision ($ALO_3$) and makes specific design choices that allowed the development of DL models that can bring improvements upon the weak supervision itself.
    \item A new method, \textit{Retrospective Beyond MeSH} (\textit{RetroBM}), for the automated development of ground-truth datasets for the large-scale evaluation of $FGSI$ predictions. 
\end{itemize}

\textit{RetroBM}, exploiting the extension of MeSH towards more fine-grained labels, developed $FGSI$ data sets with ground-truth annotations for 88 distinct fine-grained labels.
To the best of our knowledge, this is the first large-scale set of ground-truth datasets for $FGSI$, consisting of more than one million articles in total.
These data allowed us to confirm the strength of $CO$ as a heuristic for $FGSI$ of biomedical literature and proceed with the validation and evaluation experiments of this work.

\textit{DBM}, first uses the $CO$ heuristic to develop weakly labelled $FGSI$ datasets. Then, it significantly enhances these weak labels by combining $CO$ with the dictionary-based heuristics \textit{name lowercase} ($NL$) and \textit{synonyms lowercase} ($SL$), in an \textit{at-least-one} ensemble ($ALO_3$). The two heuristics examine the literal occurrence of the name and synonyms of the concept in an article.
This enhancement is particularly inserting, considering that $NL$ and $SL$ rely on the same concept terms already exploited by $CO$ itself, but use it in different ways leading to the detection of relevant articles missed by $CO$ alone. 

Finally, $DBM$ develops DL models for the $FGSI$ task without access to ground-truth labels by using the enhanced $ALO_3$ supervision and building upon the state-of-the-art approach of fine-tuning pre-trained models. 
Experiments with existing methods, namely \textit{logistic regression} ($LR$) and \textit{Coarse2Fine} ($C2F$), concluded that the $FGSI$ task is challenging and they cannot improve upon the results of the weak labels themselves. 
$DBM$ on the other hand adopts design choices to tackle the challenges of the $FGSI$ task. 
In particular, for handling the imbalance of the $FGSI$ data, $DBM$ uses the R-BCE-FL loss and under-sampling of negative instances. For avoiding overfitting on the weak supervision $DBM$ relies on early stopping considering the epoch next to the one that minimises loss ($next\_ep$). Finally, for limiting the effect of randomness it adopts majority voting among models developed with different random seeds.

The evaluation results reveal that no trained model managed to outperform the overall macro F1 ($maF1$) performance of $ALO_3$ itself, with the $DBM$ models achieving only a comparable performance.
Nevertheless, the $DBM$ models achieved competitive $FGSI$ performance overall, reaching micro F1 ($miF1$) levels above the very competitive enhanced weak labeling ($ALO_3$) itself on the new unseen labels of the evaluation experiments.
These results suggest that DL-based weakly-supervised $FGSI$ with $DBM$ is a promising direction for supporting fine-grained access to the biomedical literature. 

As regards $RetroBM$, an interesting extension would be to develop datasets for retrospective investigation of other indexing tasks, such as indexing with descriptors for groups of concepts, by adjusting the use-case selection criteria.     
As regards the improvement of the promising $DBM$ method for DL-based weakly-supervised $FGSI$, we plan to use additional label information, such as hierarchical relationships, more fine-grained text representations based on contrastive learning approaches, and probabilistic weak labels.    

\section{Acknowledgment}
This research work was supported by the Hellenic Foundation for Research and Innovation (HFRI) under the HFRI Ph.D. Fellowship grant (Fellowship Number: 697).
The data resources considered in this work, including  MeSH, MEDLINE/PubMed, and SemMedDB, were accessed courtesy of the U.S. National Library of Medicine.
We are grateful to Georgios Katsimpras for the useful discussions on Deep Learning.






\bibliography{mybibfile}

\end{document}